%% file: root.tex
\documentclass[letterpaper, 10 pt, conference]{ieeeconf}  

\IEEEoverridecommandlockouts                              

\usepackage{graphics} 
\usepackage{graphicx}
\usepackage{amsmath} 
\usepackage{amssymb}  
\usepackage{booktabs}
\usepackage{xcolor}
\let\labelindent\relax
\usepackage{enumitem} 
\usepackage[noadjust]{cite} 

\usepackage[pagebackref=true,breaklinks=true,colorlinks=true,bookmarks=false,citecolor=blue]{hyperref}
\usepackage{subfig} 
\usepackage{etoolbox}
\makeatletter
\long\def\@makecaption#1#2{%
  \vskip 10\p@
  \setbox\@tempboxa\hbox{\footnotesize #1: #2}%
  \ifdim \wd\@tempboxa >\hsize
    \footnotesize #1: #2\par
  \else
    \global \@minipagefalse
    \hb@xt@\hsize{\hfil\box\@tempboxa\hfil}%
  \fi}
\makeatother

\usepackage{multirow}
\usepackage{colortbl}
\usepackage[capitalize]{cleveref}
\definecolor{Gray}{gray}{0.93}
\definecolor{Red}{RGB}{255, 46, 23}
\definecolor{Green}{RGB}{0, 171, 79}
\definecolor{cred}{rgb}{0.85, 0.1, 0.15}
\definecolor{cgreen}{rgb}{0.25, 0.68, 0.28}
\definecolor{royalblue}{rgb}{0.25, 0.5, 0.75}
\definecolor{grayblue}{rgb}{0.9, 0.92, 0.95}

\title{\LARGE \bf
Hierarchical Skill Retrieval for Data-Efficient Adaptation of Vision-Language-Action Models
}

\author{Haoran Hao$^{1}$, Shahram Najam Syed$^{1}$, Jeff Schneider$^{1}$, Jeffrey Ichnowski$^{1}$ 
\thanks{$^{1}$Robotics Institute, Carnegie Mellon University, Pittsburgh, PA 15213.
        \{\tt\small haoranha, jeff4, jichnows\}@andrew.cmu.edu}%
}

\newcommand{\METHOD}{\underline{H}ierarchical \underline{S}kill \underline{R}etrieval}
\newcommand{\method}{HSR}

\newcommand{\Dt}{\mathcal{D}_\text{t}}
\newcommand{\Dprior}{\mathcal{D}_\text{prior}}
\newcommand{\Dret}{\mathcal{D}_\text{r}}
\newcommand{\Dlang}{\mathcal{D}_\text{lang}}
\newcommand{\Drerank}{\mathcal{D}_\text{rerank}}

\begin{document}

\maketitle
\thispagestyle{empty}
\pagestyle{empty}

\begin{abstract}

While Vision-Language-Action (VLA) models pretrained on large-scale robot datasets provide a strong foundation for robot manipulation, their performance can degrade when adapted to new tasks with limited task-specific demonstrations. Retrieval offers a practical way to reuse existing demonstrations for data-efficient adaptation, but existing methods often rely on visual similarity, state-action representations, or task-level language matching. These approaches may overlook the hierarchical structure of long-horizon manipulation tasks, where complete task matches are rare but reusable skills are often abundant.
To address this challenge, we propose {\METHOD} ({\method}), a retrieval framework for data-efficient VLA adaptation. 
Specifically, {\method} first decomposes a target task into candidate skill sequences. It evaluates each plan based on both semantic plausibility and skill reliability estimated from the prior dataset. The selected decomposition is then used for hybrid retrieval. This combines skill-level language retrieval with behavior-feature reranking to identify demonstrations that are both semantically relevant and compatible with the target task. Finally, we adapt the policy through a two-stage pretraining and finetuning pipeline, which separates general skill acquisition from task-specific adaptation. Experiments on the LIBERO benchmark and several real-world robot manipulation tasks show that {\method} improves the average success rate by 10.3\% and 21.3\% over the strongest baseline, respectively. 
These results demonstrate the effectiveness of structured skill-level retrieval for data-efficient VLA adaptation. 
Videos and code are available at \href{https://hoar012.github.io/HSR-Project/}{hoar012.github.io/HSR-Project}.
\end{abstract}

\section{Introduction}

Foundation models pretrained on large-scale datasets have shown strong generalization ability and broad transferable knowledge~\cite{clip, zhai2023sigmoid}. Motivated by recent advances in large multimodal models, vision-language-action (VLA) models \cite{kim2024openvla, black2025pi, team2024octo, o2024open, belkhale2024rt, smolvla} extend pretrained vision-language representations to robot control, aiming to transfer their general visual and linguistic understanding to downstream manipulation tasks.
A common way to adapt VLA models to downstream tasks is to finetune them on target-task demonstrations~\cite{kim2024openvla, li2025controlvla}. While effective in many settings, this approach is fundamentally limited by data scarcity: collecting high-quality robot demonstrations is expensive and time-consuming, and target datasets are often too small to fully adapt large policy models for reliable real-world deployment.

A complementary line of work improves adaptation by reusing large prior datasets as additional training data~\cite{BehaviorRetrieval, importance, COLLAGE, nasiriany2022sailor, memmel2025strap}. Among these methods, retrieval-based approaches are particularly effective because they can identify useful demonstrations from large prior datasets such as Open X-Embodiment~\cite{o2024open}.
Existing retrieval methods typically rely on similarity metrics between target data and prior data, including distances in latent state-action spaces~\cite{BehaviorRetrieval, importance} and subtrajectory-level similarity measures~\cite{nasiriany2022sailor, memmel2025strap}. 
By augmenting the target dataset with retrieved demonstrations, these methods expand the effective training set and can improve downstream task performance.

\begin{figure}[t]
  \centering
  \includegraphics[width=\linewidth]{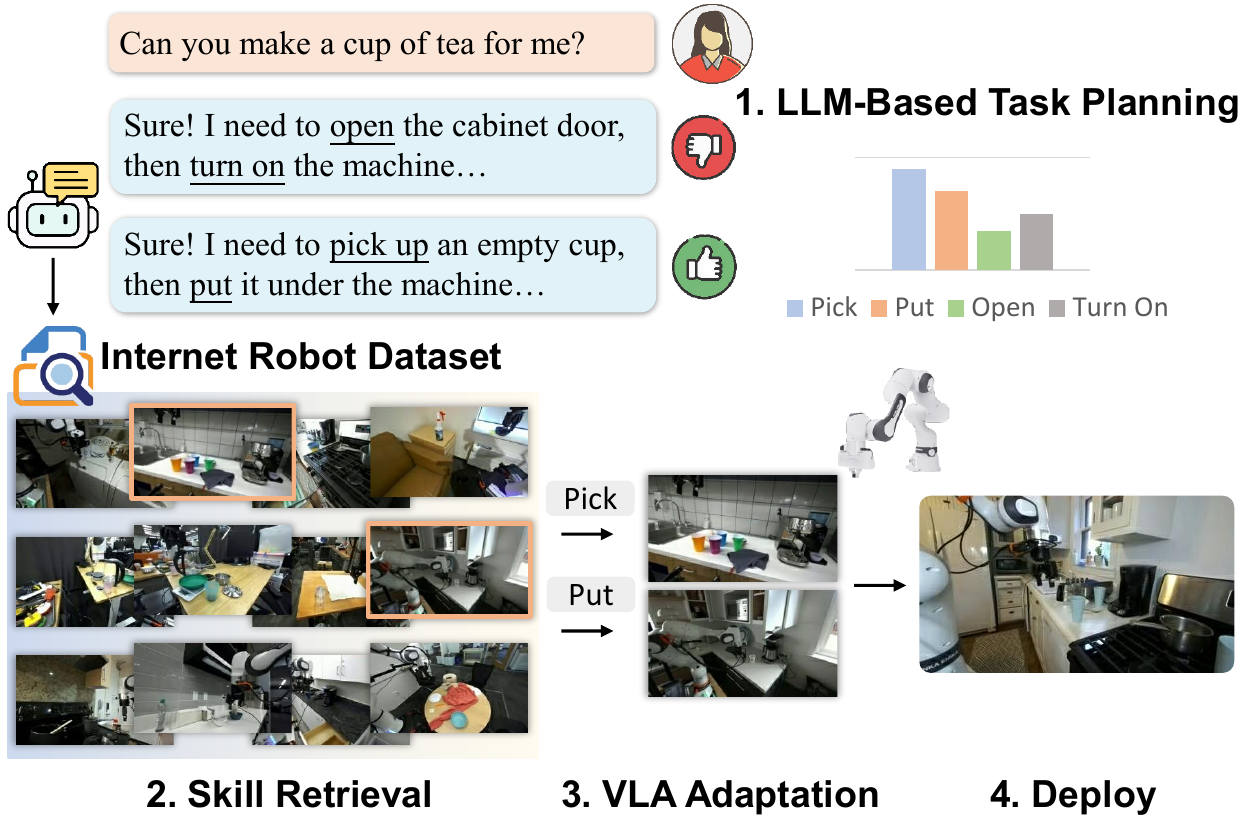}
  \vspace{-6mm}
  \caption{{\METHOD} (\method): LLM-based task decomposition enables structured demonstration retrieval for data-efficient VLA adaptation.}
  \label{fig:teaser}
  \vspace{-4mm}
\end{figure}

Although retrieval-based methods have shown promising results in several settings, most existing approaches are designed for vision-based policies and primarily rely on visual or motion similarity. As a result, they often overlook higher-level task semantics and relationships between tasks, and thus do not fully exploit language instructions.
VLA models, on the other hand, have demonstrated strong capabilities in following language instructions \cite{shi2025hi, yang2025instructvla} and transferring knowledge across related tasks \cite{black2025pi,LBM, black2024pi_0}, such as manipulating different objects using similar actions. This suggests that language can provide an important semantic signal for improving task generalization.

However, language-based retrieval is also limited when it relies only on coarse instruction similarity. For complex long-horizon tasks, such as \emph{``make a cup of tea''}, directly retrieving data using the full task instruction is often difficult, since complete task matches rarely exist in prior datasets. 
In contrast, many of the underlying skills required to complete such tasks, such as \emph{``grasp a teapot''} or \emph{``pick up a tea bag''}, are common and well represented. 
Effectively leveraging the hierarchical structure of complex tasks and retrieving data at the level of reusable skills therefore remains an open challenge.

To address these issues, we propose {\METHOD} ({\method}), a framework for data-efficient VLA adaptation based on task decomposition and skill-level retrieval. Specifically, {(1)} we first cluster the prior dataset into reusable skills, such as Pick, Open, and Close. Given a target task, we use a Large Language Model (LLM)-based planner to decompose the task into subtasks grounded in the available skills. {(2)} We then score candidate plans, select the most suitable decomposition, and use the resulting subtasks as queries for retrieval.
{(3)} Finally, we adapt the VLA model in two stages: first by pretraining on language-retrieved skill data, and then by finetuning on the target demonstrations together with feature-reranked retrieved samples.

The \method{} framework leverages both the structure of the target task and the prior dataset, enabling data-efficient adaptation of pretrained VLA models under limited target data. We evaluate the proposed approach in the LIBERO~\cite{libero} simulation environment and on several real-world long-horizon robot manipulation tasks. Experimental results show that {\method} achieves the best average performance across a wide range of tasks and provides particularly large gains on long-horizon composite manipulation tasks. Notably, {\method} improves the success rate by 10.3\% on LIBERO and by 21.3\% on real-world tasks relative to the strongest baseline.

Our contributions are summarized as follows:

\begin{itemize}
\item We propose a hierarchical retrieval framework for long-horizon manipulation, which decomposes tasks into skill sequences and retrieves demonstrations at the skill level based on task structure and semantics.
\item We introduce a two-stage adaptation strategy. It first uses retrieved demonstrations to acquire transferable skills, and then finetunes on target data together with feature-reranked retrieved samples.
\item We validate the proposed framework on both the LIBERO benchmark and real-world long-horizon manipulation tasks, showing improved success rates and data efficiency over strong retrieval-based baselines.
\end{itemize}

\section{Related Work}

\subsection{Vision-Language Conditioned Manipulation}
Earlier work on robot manipulation using deep neural networks typically restricted the input modality to either visual observations or robot states~\cite{billard2019trends}. In recent years, vision-language models (VLMs)~\cite{clip, zhai2023sigmoid, llava} pretrained on large-scale image-text pairs have demonstrated strong capabilities in multimodal perception and understanding. Based on these multimodal foundation models, Vision-Language-Action (VLA) policy models~\cite{kim2024openvla, team2024octo, o2024open, belkhale2024rt, smolvla, qu2025spatialvla,oc-vla,brohan2022rt} aim to transfer the general knowledge learned by VLMs to robot control. By training on multi-task datasets and conditioning on both language instructions and visual observations, VLA models show improved generalization in vision-language manipulation tasks. Pretrained VLA models can leverage large-scale datasets such as Open X-Embodiment~\cite{o2024open}, enabling cross-task generalization and improved instruction understanding in robot manipulation. However, despite containing rich general knowledge, VLA models often perform suboptimally on specific downstream tasks. In many practical scenarios, downstream tasks lack sufficient data for training or finetuning. As a result, how to effectively adapt VLA models to data-scarce downstream tasks remains an open and underexplored problem.

\subsection{Retrieval for Model Adaptation}
Retrieval-based methods aim to select relevant data from existing datasets, such as DROID~\cite{khazatsky2024droid} and Open X-Embodiment~\cite{o2024open}, to improve adaptation to specific target tasks \cite{Hao_2025_CVPR}. Most prior work learns latent representations of trajectories and performs similarity-based retrieval. For example, some methods encode state-action pairs~\cite{BehaviorRetrieval}, optical flow~\cite{lin2024flowretrieval}, or sub-trajectories~\cite{nasiriany2022sailor, memmel2025strap}, and retrieve data based on visual or motion similarity. Other approaches extend this idea by incorporating multiple modalities~\cite{COLLAGE} or applying importance weighting between prior and target datasets~\cite{importance}. There are also methods that retrieve relevant past experiences and directly learn from them~\cite{expresvla}.
However, these approaches mainly rely on visual or motion-level similarity, and often do not explicitly consider the semantic and hierarchical structure of manipulation tasks, which is crucial for modern VLA models. Trajectories with similar motion patterns may correspond to different skills, such as pushing versus picking, leading to ambiguous or less relevant retrieval results.
Recent work has started to explore semantic-level reasoning in robot tasks. DROC~\cite{DROC} shows that LLMs can generate language-based corrections, which are then used as queries to retrieve relevant knowledge from a knowledge base. MT3~\cite{athousand} decomposes tasks into alignment and interaction policies, enabling task composition and improved generalization through test-time retrieval. Motivated by these insights, we propose a framework that decomposes long-horizon manipulation tasks into subtasks and retrieves semantically related experiences from a prior dataset.

\subsection{Robot Task Planning}

Task planning aims to decompose and schedule complex tasks by combining symbolic reasoning, motion planning, and learning-based methods~\cite{Hierarchical}. Effective planning enables zero-shot generalization to novel tasks by reusing existing skills~\cite{sahni2017learning, dalal2024manipgen}, reducing the need for additional data to learn composite tasks.
Some approaches rely on human-defined transition models or learn such models from data to generate composed plans~\cite{learning-composable, liulearning, mao2022pdsketch, abil}. However, building such models often requires substantial human effort, large training datasets, and additional components, such as object detectors~\cite{mao2022pdsketch}.
More recent work explores the reasoning capabilities of LLMs for task planning~\cite{team2025gemini, jin2024robotgpt, li2024embodied, li2024manipllm, saycan,vlmguidedmanipulation}. The resulting subtasks are typically executed by separate policy modules~\cite{li2024manipllm, singh2023progprompt}. Most existing approaches focus on task decomposition at test time, while placing less emphasis on how the required skills are trained or adapted.
In this work, we use task decomposition as a training-time retrieval interface. Given a target task, we identify a sequence of semantically relevant skills and retrieve training demonstrations that support each skill, thereby improving data-efficient adaptation of the policy.

\begin{figure*}[ht]
  \centering
  \includegraphics[width=0.92\linewidth]{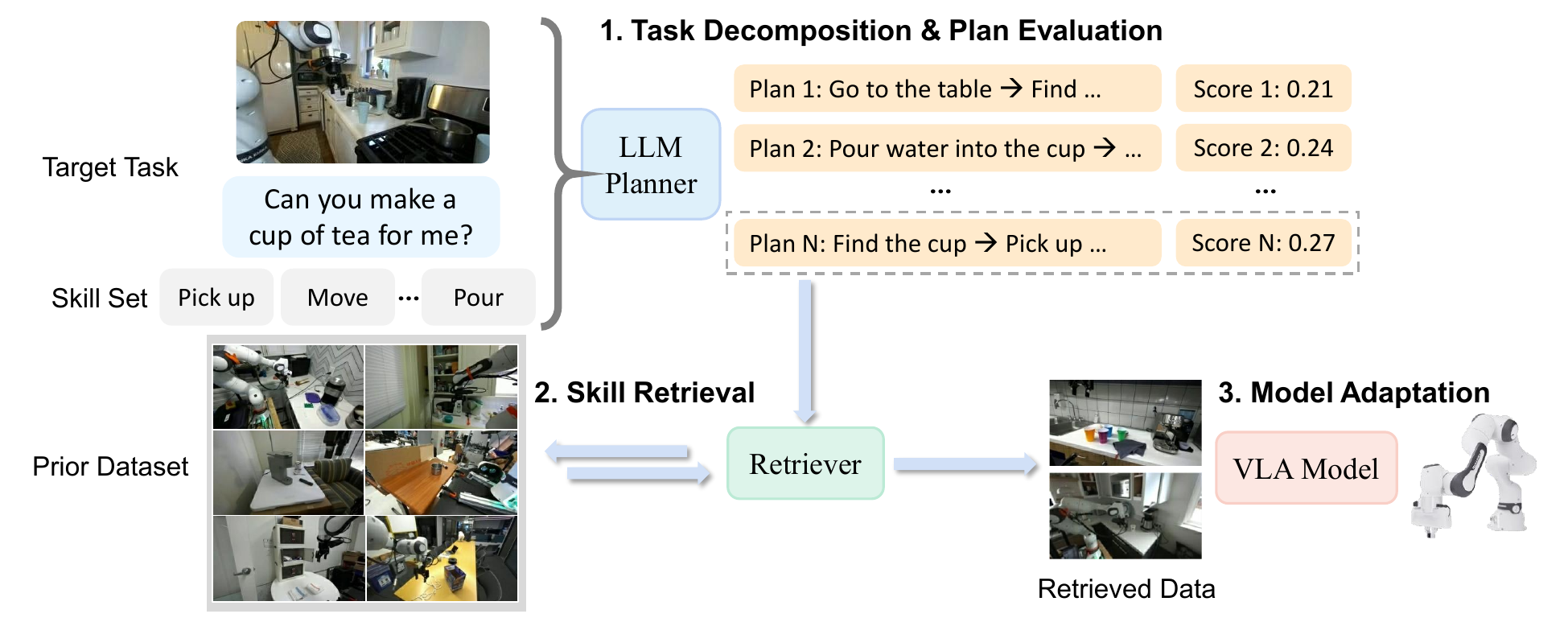}
  \vspace{-3mm}
  \caption{\textbf{Overview of the Proposed {\method} Framework}. HSR identifies existing skills in the prior dataset based on task-instruction clustering. Given a long-horizon target task, \emph{1.} HSR first decomposes the task into several subtasks, evaluates and selects the plan with the highest score; \emph{2.} for each subtask, HSR retrieves data based on the subtask instruction, and further reranks using representative demonstrations from the target task to filter out low-consistency data; \emph{3.} HSR first pretrains the policy on retrieved data to learn general skills, and then finetunes it on task-related data to adapt to the target task.}
  \label{fig:framework}
\end{figure*}

\begin{figure}[t]
  \centering
  \includegraphics[width=0.95\linewidth]{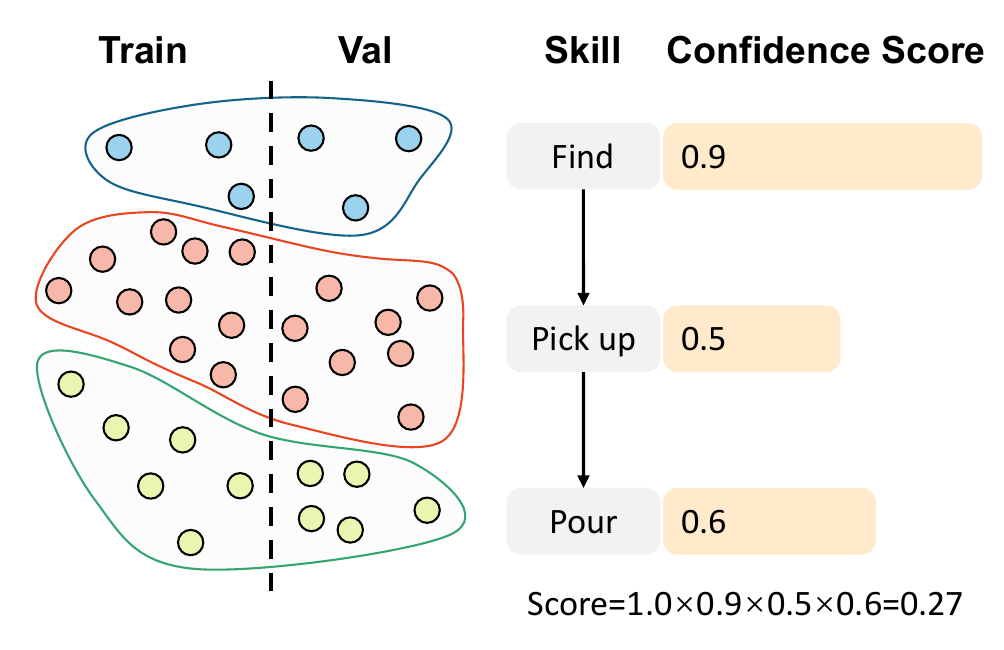}
  \vspace{-2mm}
  \caption{\textbf{Illustration of Plan Evaluation}. We first cluster $\Dprior$ into reusable skills. Each clustered skill is then evaluated by finetuning a pretrained VLA model on the training split and measuring the BC loss on its validation set. The resulting loss serves as a lightweight proxy for skill reliability, enabling efficient plan evaluation without costly environment rollouts.}
  \label{fig:score}
  \vspace{-4mm}
\end{figure}

\section{Problem Statement}
We study few-shot adaptation of a pretrained vision-language-conditioned policy using a small set of target demonstrations together with a large prior dataset. Given a pretrained VLA policy $\pi_\theta$, the policy maps a visual observation $o_t$ and a language instruction $T$ to an action $a_t$: $a_t = \pi_\theta(o_t, T)$. For each target task, we are given a target dataset $\Dt = \{\tau_i\}_{i=1}^{N}$, where each trajectory $\tau_i = \{(o_t, a_t)\}_{t=1}^{H}$ is an expert demonstration. We assume access to a large prior dataset $\Dprior$, which contains demonstrations from diverse tasks and domains. Our objective is to identify a useful subset $\Dret \subset \Dprior$ such that adaptation on $\Dt \cup \Dret$ yields a policy that better matches expert behavior and performs well on the target task.

\section{Method}
\label{sec:framework}
\subsection{Hierarchical Task Decomposition}
Language-based retrieval is often insufficient for robot manipulation tasks, where high-level instructions are ambiguous and rarely match complete demonstrations in the prior dataset. 
This limitation is particularly severe for long-horizon and compositional tasks, such as \emph{``make a cup of tea''}, which are seldom observed as monolithic behaviors in practice. 
However, these tasks are typically composed of reusable skills that recur across different tasks, such as \emph{``pick up a cup''} and \emph{``turn on a machine''}. Motivated by this observation, we decompose a target task into a sequence of semantic subtasks and use each subtask as a query to retrieve relevant demonstrations from $\Dprior$. This skill-level retrieval enables more precise reuse of prior experiences and improves adaptation to long-horizon manipulation. The overall {\method} framework is illustrated in Fig.~\ref{fig:framework}.

Although LLMs have shown strong performance in task planning, prior work such as SayCan~\cite{saycan} points out an important limitation: LLMs are not grounded in the robot's actual capabilities and may therefore generate plans containing subtasks that are difficult or impossible to execute. As a result, purely language-based planning may produce decompositions that are poorly matched to the policy's learned skills.
To mitigate this issue, we leverage $\Dprior$ to estimate the reliability of different skills and use this information to guide task decomposition for retrieval. Our key idea is to prefer decompositions composed of skills that are both semantically relevant to the target task and well supported by the available data.
Specifically, we first extract text embeddings for all task instructions in $\Dprior$ and apply K-means clustering to group them into a set $S = \{s_1, \ldots, s_K\}$, where each cluster $s_i$ corresponds to a semantically coherent skill, such as \emph{Find}, \emph{Pick up}, or \emph{Pour}.

For each $s_i$, we estimate its reliability using the pretrained VLA model.
Rollout-based evaluation provides the most direct measure of policy performance, but is computationally expensive and often impractical, especially in real-world settings. Instead, following DataMIL~\cite{dass2025datamil}, we use held-out behavior cloning (BC) loss as a lightweight proxy for skill reliability. As illustrated in Fig.~\ref{fig:score}, we split $\Dprior$ into training and validation sets and finetune the pretrained VLA model on the training split to obtain $\pi_{\theta'}$. We then evaluate $\pi_{\theta'}$ on the validation trajectories associated with each skill cluster.
For a validation trajectory $\tau=\{(o_t,a_t)\}_{t=1}^{H_\tau}$, let $\ell_{\mathrm{BC}}(\pi_{\theta'}(o_t,T), a_t)$ denote the per-step BC loss. We define the average trajectory-level BC loss as
\begin{equation}
\mathcal{L}_{\mathrm{traj}}(\tau)
=
\frac{1}{H_\tau}
\sum_{t=1}^{H_\tau}
\ell_{\mathrm{BC}}(\pi_{\theta'}(o_t,T), a_t).
\end{equation}
For a skill cluster $s_i$, let $\mathcal{V}_i$ denote its validation trajectories. The validation loss of $s_i$ is computed as
\begin{equation}
\mathcal{L}_{\mathrm{val}}(s_i)
=
\frac{1}{|\mathcal{V}_i|}
\sum_{\tau\in\mathcal{V}_i}
\mathcal{L}_{\mathrm{traj}}(\tau),
\end{equation}
and $H(s_i)$ denotes the average trajectory length in $\mathcal{V}_i$.
To ensure comparability across skill clusters, we normalize both quantities using min-max normalization.
Since long-horizon skills are more susceptible to compounding errors, we define the \emph{Skill Confidence Score} as
\begin{equation}
C(s_i) =
\exp\left(
-\alpha \widetilde{\mathcal{L}}_{\mathrm{val}}(s_i)
-\beta \widetilde{H}(s_i)
\right),
\end{equation}
where $\widetilde{\mathcal{L}}_{\mathrm{val}}(s_i)$ and $\widetilde{H}(s_i)$ denote the normalized validation loss and average trajectory length, respectively. 
We use fixed weights $\alpha=0.6$ and $\beta=0.8$ across all experiments.
Rather than estimating the true success probability, $C(s_i)$ serves as a rollout-free proxy for skill reliability, favoring skills that are easier to imitate and less susceptible to error accumulation during long-horizon execution.

We then query the LLM with the target task description and the set of available skills to generate candidate decomposed plans. Each plan is represented as an ordered sequence of subtasks $p = (u_1, u_2, \ldots, u_n)$. For each subtask $u_j$, we assign its nearest skill cluster based on text-embedding similarity, denoted as $s(u_j)$. In addition, we prompt the LLM to provide a semantic score for each plan, reflecting its task coverage, logical consistency, and semantic plausibility with respect to the target instruction. The score is normalized to $S_{\mathrm{sem}}(p) \in (0,1]$ before plan ranking.
We then evaluate each candidate plan by combining the semantic score with the confidence scores of its constituent skills:
\begin{equation}
\operatorname{Score}(p) =S_{\mathrm{sem}}(p)\cdot\prod_{j=1}^{n} C\big(s(u_j)\big).
\end{equation}
We select the plan with the highest score as the final decomposition, and use it for subsequent retrieval from $\Dprior$.

\subsection{Hybrid Skill Retrieval}
Language-based retrieval is effective at identifying demonstrations that are semantically relevant to a target task. However, VLA policies are also sensitive to low-level factors such as environment layouts, lighting conditions, and camera viewpoints. 
As a result, demonstrations that are similar in language instruction may still differ substantially from the current execution context, which can degrade policy adaptation and execution.
To address this issue, we propose a hybrid skill retrieval strategy that combines language-based retrieval with behavior-feature reranking. The key idea is to first use language similarity to ensure semantic relevance, and then use behavior similarity to select demonstrations that are more compatible with the target task.

Specifically, given a subtask, we first retrieve the top $M\%$ of episodes from $\Dprior$ according to language similarity.
We then flatten these episodes into candidate frames and rerank them using behavior features. Following Behavior Retrieval~\cite{BehaviorRetrieval}, we train a VAE encoder on $\Dprior$ to extract their behavior features.
Let $z_f$ denote the latent feature of a candidate frame $f$, and let $\{z_j^t\}_{j=1}^{N_t}$ denote the latent features extracted from $\Dt$, where $N_t$ is the number of target frames. The behavior similarity score of $f$ is defined as
\begin{equation}
s(f) = \max_{1 \le j \le N_t} \left( -\| z_f - z_j^t \|_2 \right),
\end{equation}
which corresponds to the negative distance to the nearest target example in latent space.
We retain the top $F\%$ of candidate frames with the highest similarity scores as the final retrieved data for policy adaptation. By combining semantic alignment at the episode level with behavior compatibility at the frame level, this hybrid retrieval strategy improves the relevance and effectiveness of the retrieved skills under varying environmental conditions.

\subsection{Policy Adaptation}
To better exploit the retrieved data, we adopt a two-stage policy adaptation scheme. Let $\Dlang$ denote the training samples from the language-retrieved episodes and let $\Drerank$ denote the reranked frames constructed from $\Dlang$.
In the first stage, we pretrain the policy on $\Dlang$ to acquire reusable behaviors. In the second stage, we finetune the policy on $\Dt \cup \Drerank$ for task-specific adaptation.
This design differs from prior single-stage co-training methods~\cite{BehaviorRetrieval, importance}, which jointly optimize the policy on the retrieved data $\Dret$ and the target data $\Dt$ in one step. In contrast, our two-stage strategy explicitly separates general skill acquisition from task-specific adaptation and avoids treating all retrieved data as equally informative.

\input{table/main}
\section{Experiments}
\textbf{Models and Training.}
We use the pretrained SmolVLA-450M~\cite{smolvla} model as the base policy model. During adaptation, we update only the parameters of the action expert, resulting in approximately 100M trainable parameters. In the {\method} framework, we employ Qwen3-VL-4B~\cite{bai2025qwen3} as the LLM planner for task decomposition. We use $K=10$ clusters for both simulation and real-world experiments.
For each target task, the planner generates 10 candidate plans, and we select the plan with the highest score.
We use a pretrained BERT model~\cite{devlin2019bert} to extract text embeddings for retrieval.

\textbf{Environments.}
We evaluate the proposed framework in both simulation and real-world settings. In simulation, we use the LIBERO benchmark~\cite{libero}. For real-world evaluation, we design long-horizon manipulation tasks executed on a 7-DoF xArm robot. All experiments are conducted using a single Nvidia RTX 5090 GPU.

\textbf{Simulation.} We evaluate on the LIBERO long-horizon subset, which consists of 10 composite manipulation tasks requiring diverse skills such as pick-and-place and drawer closing. For each task, we construct a target dataset $\Dt$ containing 5 demonstrations. Following \cite{importance, memmel2025strap}, we use LIBERO-90 as $\Dprior$, which includes 90 tasks with 50 demonstrations per task. 
We retrieve the top 10\% of episodes from $\Dprior$ in the first stage and retain the top 30\% of these candidates after reranking in the second stage, corresponding approximately to 3\% of frames from $\Dprior$. We evaluate the learned policy over 50 episodes using 3 random seeds and report the average success rate with standard deviation.

\begin{figure*}[th]
  \centering
  \subfloat[Drawer]
  {
\label{grounding:subfig1}\includegraphics[width=0.21\linewidth]{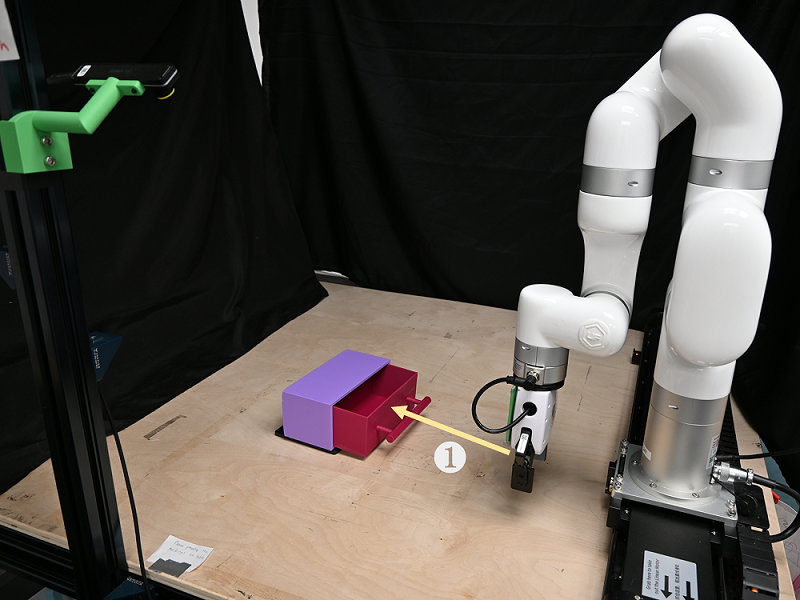}
  }
  \subfloat[Trashcan]
  {
\label{grounding:subfig2}\includegraphics[width=0.21\linewidth]{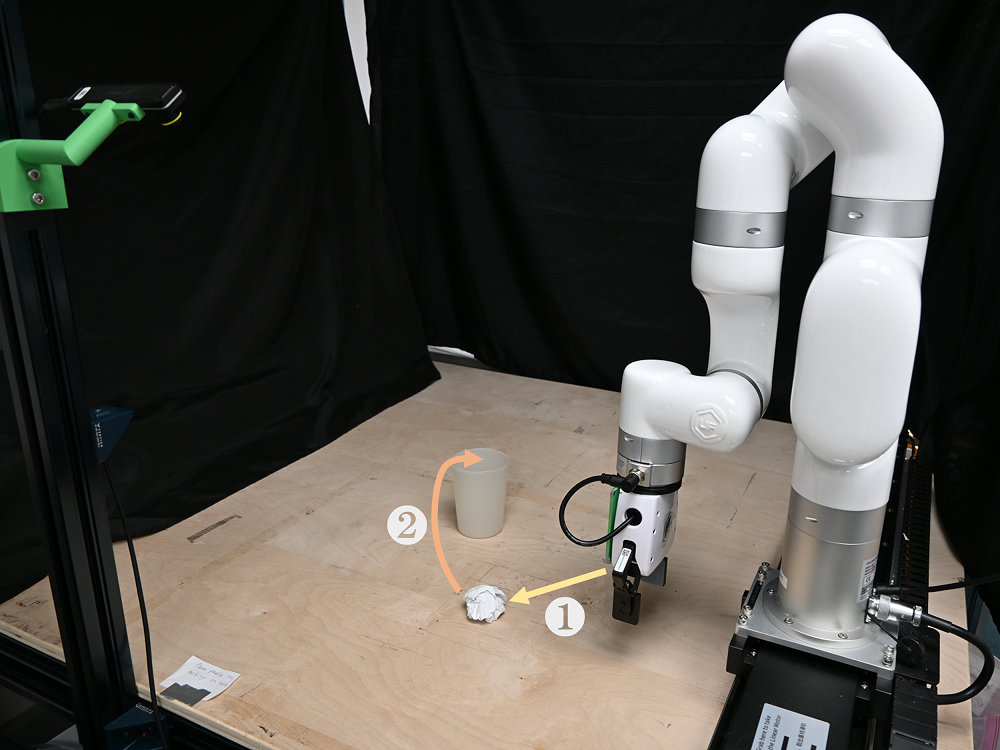}
  }
   \subfloat[Cup-Drawer]
  {
\label{grounding:subfig3}\includegraphics[width=0.21\linewidth]{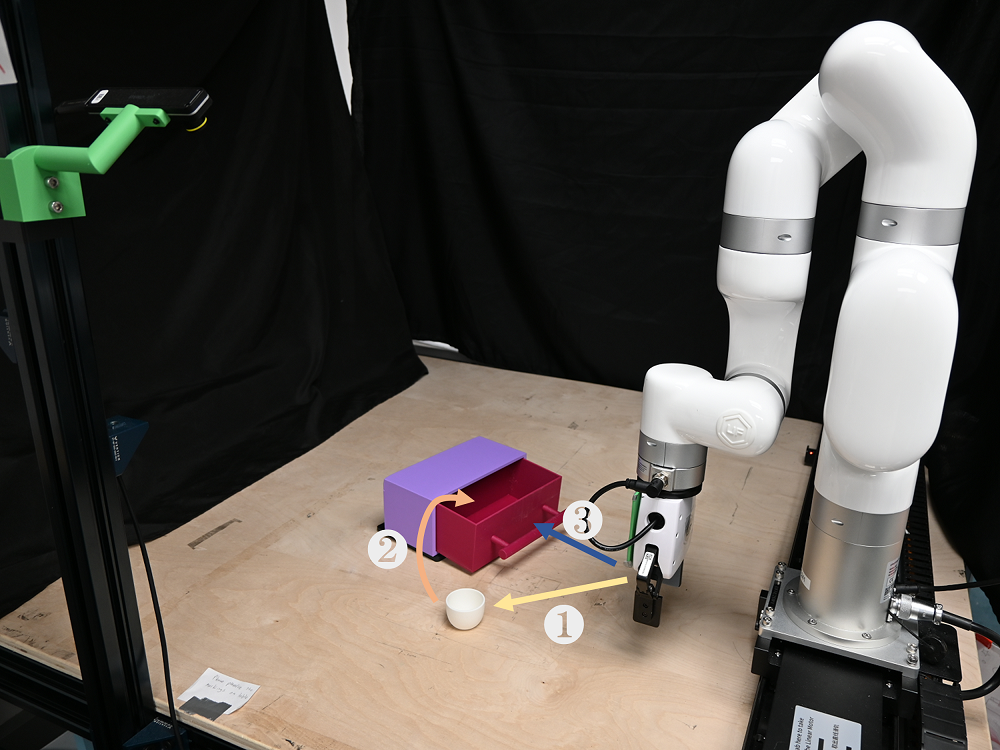}
  }
  \subfloat[Cup-Tea]
  {
\label{grounding:subfig4}\includegraphics[width=0.21\linewidth]{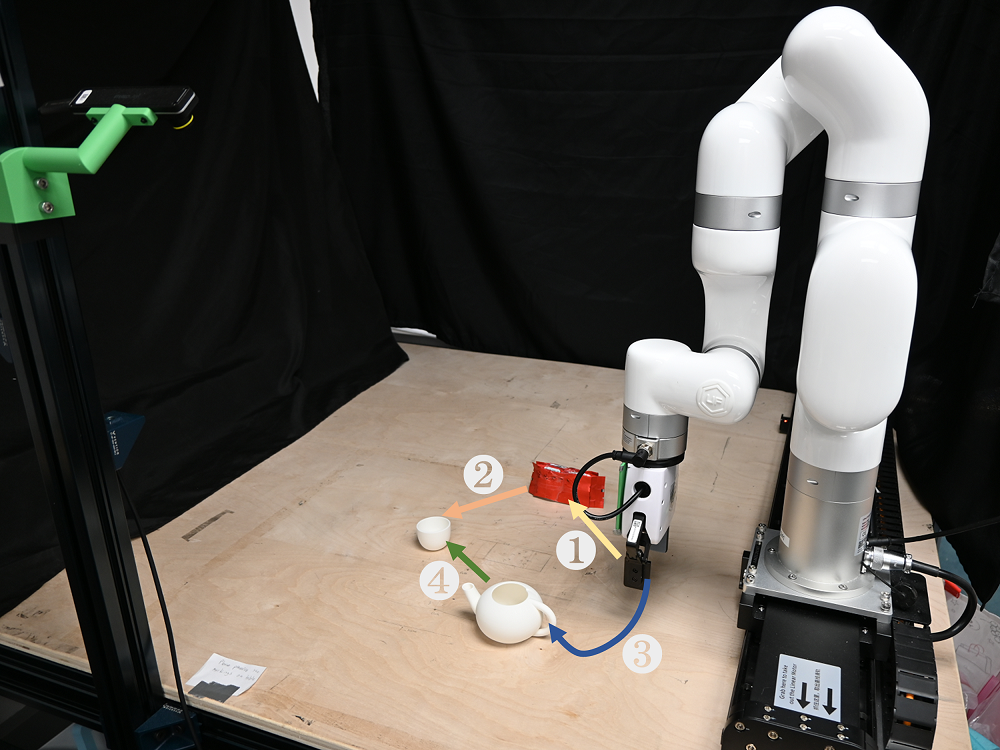}
  }
  \caption{\textbf{Experimental Setup of Real-robot Tasks}. We evaluate using a 7-DoF xArm on several manipulation tasks: \textbf{(a) Drawer}: Close the drawer, \textbf{(b) Trashcan}: Pick up the crumpled paper and throw it in the trash can, \textbf{(c) Cup-Drawer}: Put the cup into the drawer and close it, \textbf{(d) Cup-Tea}: Put the teabag next to the cup and pour water into the cup.} 
  \label{fig:setting}
  \vspace{-3mm}
\end{figure*}

\textbf{Real-World.}
We design several real-world manipulation tasks with varying levels of complexity, including \emph{Drawer}, \emph{Trashcan}, \emph{Cup-Drawer}, and \emph{Cup-Tea} (Fig.~\ref{fig:setting}). For each task, we collect 20 demonstrations as $\Dt$. We randomly sample 10k trajectories from the DROID dataset~\cite{khazatsky2024droid} as $\Dprior$. In the first stage, we use the reranked top 3\% of frames from $\Dprior$ for skill pretraining. In the second stage, we finetune the policy only on the target dataset $\Dt$, as the cross-embodiment gap between the source data and our xArm platform makes direct co-training less effective.
We evaluate the learned policy over 20 trials per task and report the success rate. 
Each trial uses a different initial state, while the set of states is identical across models for fair comparison.

\textbf{Baselines}.
We compare \method{} with baselines covering target-only finetuning, naive data mixing, language-based retrieval, optical-flow retrieval, subtrajectory retrieval, and state-action representation-based retrieval.

\begin{description}[
    leftmargin=5pt,
    labelindent=0pt,
    itemindent=0pt,
    labelsep=0.5em,
    font=\normalfont\bfseries
]
\item[Behavior Cloning (BC)] finetunes the VLA model using only the target dataset $\Dt$.
\item[Random] randomly samples 10\% of the data from the prior dataset $\Dprior$ and co-trains the VLA model with the target dataset $\Dt$.
\item[All Data] co-trains the VLA model on the entire prior dataset $\Dprior$ together with the target dataset $\Dt$.
\item[Language] performs simple language-similarity-based retrieval and co-trains the VLA model on the retrieved dataset $\Dret$ and the target dataset $\Dt$.
\item[Flow Retrieval (FR)~\cite{lin2024flowretrieval}] retrieves data based on similarity computed from samples' optical flows.
\item[STRAP~\cite{memmel2025strap}] employs vision foundation models and dynamic time warping to retrieve sub-trajectories from $\Dprior$.
\item[Behavior Retrieval (BR)~\cite{BehaviorRetrieval}] trains a VAE to encode state-action pairs and retrieves data from $\Dprior$ based on similarity in the learned latent space.
\item[Importance Weighted Retrieval (IWR)~\cite{importance}] uses the same VAE encoder but retrieves data from $\Dprior$ based on estimated importance weights.
\end{description}

\subsection{Simulation Experiments}
Table~\ref{tab:baselines_sim} summarizes the results on the LIBERO benchmark. Finetuning on the target dataset alone (BC) achieves limited performance due to the small number of demonstrations. 
Random and All Data achieve only modest gains despite using additional prior data, suggesting that simply increasing data volume is insufficient when the prior dataset contains redundant or task-irrelevant demonstrations.

Language-based retrieval improves over random sampling, but remains limited on tasks with more complex instructions, suggesting that language similarity alone is insufficient to capture fine-grained semantics. State-action-based retrieval methods, including BR and IWR, further improve performance by leveraging trajectory-level similarity. However, their gains are smaller on composite manipulation tasks, where higher-level task structure matters more.
STRAP is the closest baseline to {\method}, as it also performs retrieval at the subtrajectory level. Nevertheless, its vision-based similarity criterion is less effective at capturing the semantic structure of long-horizon tasks, which limits its performance on tasks that require reusable skills.

Overall, {\method} retrieves more useful demonstrations and achieves the best average performance. On tasks that mainly require a single \emph{pick-and-place} skill, {\method} performs comparably to BR, indicating that directly retrieving highly similar demonstrations can already be sufficient in simpler settings. In contrast, the advantage of {\method} becomes more evident on complex composite tasks. For example, on Mug-M, {\method} outperforms the strongest competing method on this task, STRAP, by 27.3\%. On the challenging Bowl-C task, it is the only method that improves over pure behavior cloning. These results demonstrate the effectiveness of {\method} in selecting useful prior data and enabling effective policy adaptation.

\subsection{Real-World Experiments}
For real-robot tasks, data scarcity poses a significant challenge. As shown in \cref{tab:real}, BC trained only on target demonstrations achieves a success rate below 40\,\%. 
BR and IWR provide only limited improvement, suggesting that directly transferring prior data remains challenging under the substantial cross-embodiment gap between the source data and the target tasks.
In contrast, {\method} improves performance by retrieving relevant demonstrations in a structured manner. As a result, {\method} achieves higher success rates across real-robot tasks of varying horizons. Notably, the advantage of {\method} becomes more pronounced on long-horizon tasks. For the \emph{``make a cup of tea''} task, which requires several hundred low-level actions to complete, BC trained only on the target demonstrations succeeds in 3 out of 20 trials, whereas {\method} achieves 9 out of 20 successful executions, demonstrating a substantial improvement on complex multi-step manipulation.

\input{table/real}

\subsection{Ablation Studies}

\begin{figure}[t]
  \centering
  \includegraphics[width=\linewidth]{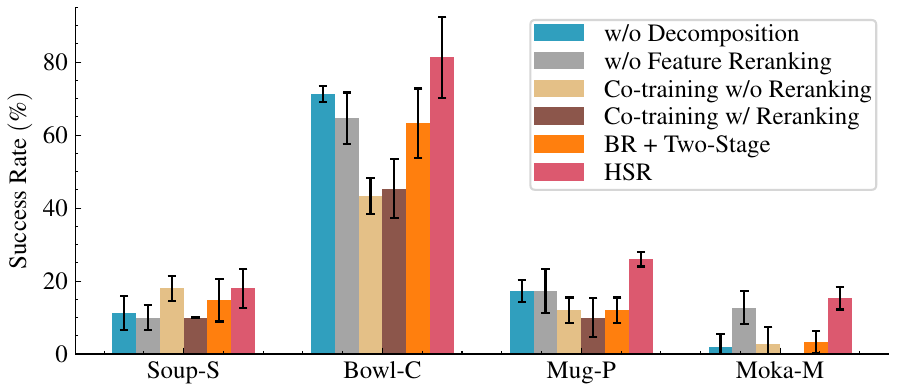}
  \vspace{-6mm}
  \caption{\textbf{Results of Ablation Studies.} We conduct ablation studies on four LIBERO tasks to evaluate the effectiveness of key components of {\method}, including task decomposition, feature reranking, and the pretraining-finetuning strategy. The results show that each component improves performance, with the full {\method} framework achieving the highest success rates across all tasks.}
  \vspace{-4mm}
  \label{fig:ablation}
\end{figure}

\textbf{Task Decomposition.}
We remove the task decomposition module and directly retrieve data using the original task instruction, while keeping the rest of the framework unchanged. As shown in Fig.~\ref{fig:ablation}, performance drops noticeably, especially on tasks with more complex instructions such as Mug-P and Moka-M. This suggests that using the full task description alone is often insufficient to retrieve demonstrations required by long-horizon manipulation. In contrast, task decomposition breaks a complex instruction into semantically meaningful subtasks, enabling more precise retrieval and improving overall performance.

We further compare different decomposition strategies in Table~\ref{tab:ablation:decomposition}, which reports the average success rate across all LIBERO-10 tasks. Raw LLM-generated plans already achieve competitive performance, indicating that LLMs can produce reasonable high-level decompositions. Incorporating the available skill set into the prompt further improves performance, suggesting that grounding the planner with executable skills leads to better task decomposition. Adding the plan evaluation module of {\method} yields a further improvement, demonstrating the value of selecting decompositions that are both semantically relevant and well supported by the prior data. Manually selected plans achieve performance comparable to that of {\method}, indicating that the proposed plan evaluation strategy can approach the quality of manually designed decompositions.

\input{table/ablation}

\begin{figure*}[t]
  \centering
  \includegraphics[width=0.9\linewidth]{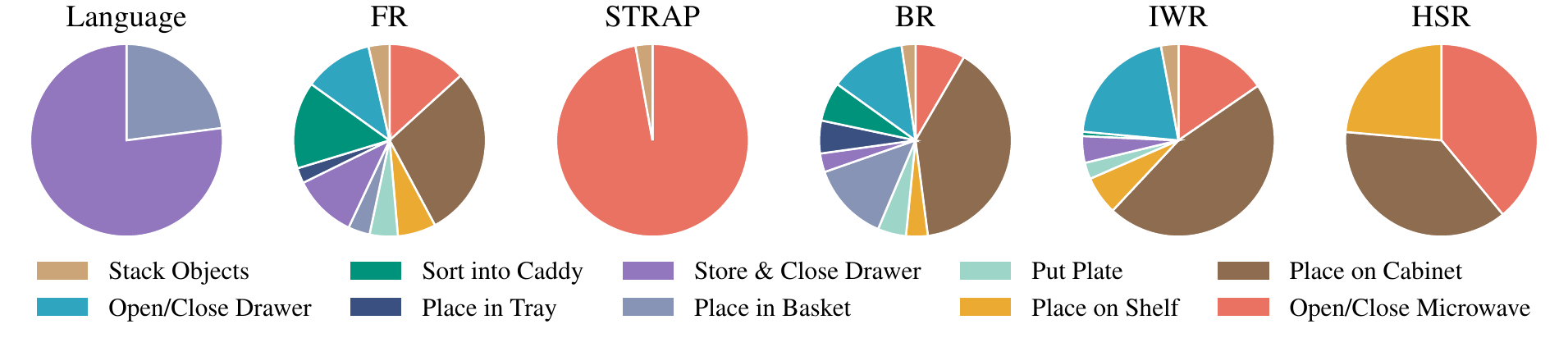}
  \vspace{-2mm}
  \caption{\textbf{Comparison of Retrieved Data.} Distribution of retrieved data for the task \emph{``put the yellow and white mug in the microwave and close it''}. \method{} effectively retrieves demonstrations across subtasks covering both object-placing and microwave-closing skills.}
  \label{fig:data}
  \vspace{-2mm}
\end{figure*}
\textbf{Feature Reranking.}
As shown in Fig.~\ref{fig:ablation}, removing the feature reranking module and using only language-retrieved data leads to a clear performance drop, especially on tasks that require precise alignment with the target execution context, such as Soup-S. This suggests that language similarity alone is insufficient to capture low-level compatibility between the retrieved demonstrations and the target task. In contrast, the feature reranking module filters out demonstrations that are less compatible in behavior space and retains those that better match the target task, thereby improving policy adaptation.

We also ablate the retention ratio after feature reranking, with results shown in Fig.~\ref{fig:ablation:percentage}. The results indicate that retaining a moderate fraction of the reranked data yields the best performance on most tasks. Retaining too much data introduces additional noise, while retaining too little data may discard useful demonstrations. 

\textbf{Training method.}
Another way to leverage retrieved data is to co-train the policy on both the target and retrieved datasets in a single stage. We compare this co-training baseline with our proposed two-stage pretraining-finetuning strategy. As shown in Fig.~\ref{fig:ablation}, both co-training variants, with and without feature reranking, achieve lower success rates on most tasks.
One possible reason is that retrieved demonstrations are typically larger in scale, noisier, and less well aligned with the target task than the target demonstrations. Co-training may therefore bias optimization toward the retrieved data, which can be suboptimal when the gap between the retrieved and target datasets is large. In contrast, our strategy first uses retrieved data to learn general skills and then finetunes on the most relevant demonstrations, which helps mitigate this issue and provides greater robustness to diverse data sources and embodiments.

We further evaluate two-stage adaptation by replacing our retrieval method with BR. This combined approach improves over the original BR baseline, but still underperforms {\method}, further demonstrating the benefit of our retrieval strategy.

\subsection{Retrieved Data Study}
To illustrate the effect of different retrieval methods, we consider a representative task from the LIBERO benchmark, \emph{``put the yellow and white mug in the microwave and close it.''} The retrieved demonstrations are visualized in Fig.~\ref{fig:data}. 
The language-based retrieval baseline relies only on the original task description and may miss task-specific nuances. FR, STRAP, BR, and IWR incorporate additional motion- or behavior-level information, but may still retrieve demonstrations from tasks with similar motions and different goals. 
Such mismatched data can mislead policy learning.
In contrast, {\method} decomposes the long-horizon task into subtasks and retrieves demonstrations for each subtask independently. This allows the method to retrieve demonstrations that are more closely aligned with the specific requirements of each step, leading to more task-relevant retrieved data and improved performance.

\section{Conclusion}
We present {\METHOD} ({\method}), a framework for data-efficient adaptation of VLA models to downstream manipulation tasks. By exploiting the hierarchical structure of long-horizon tasks, {\method} combines high-level semantic decomposition with low-level behavior-aware retrieval to identify useful prior demonstrations. This allows the policy to reuse transferable skills from prior datasets, rather than relying on exact task-level matches. Experiments on the LIBERO benchmark and real-world long-horizon manipulation tasks show that {\method} consistently improves adaptation performance under limited target data.

\textbf{Discussion and Limitations.} Despite these promising results, several challenges remain. 
First, the current framework is evaluated only on manipulation tasks and a single VLA backbone. Evaluating {\method} across different policy architectures, robot embodiments, and data modalities, such as videos or images collected in diverse domains, remains an important direction for improving robustness and data efficiency.
Second, although we use an LLM for task decomposition, tighter integration of language models into closed-loop control, such as online replanning and failure recovery, remains an important direction for future work.
In addition, even VLAs trained with additional data may still fail during deployment. Developing mechanisms such as test-time adaptation \cite{hao2026farfailureawareretrytesttime} to handle such failures is also a promising direction for future work.

\bibliographystyle{IEEEtran}
\bibliography{IEEEabrv,root}

\end{document}

%% file: table/main.tex
\begin{table*}[t]
    \caption{\textbf{Comparison on LIBERO Benchmark}. We compare our method with behavior cloning (BC), training with all data, language-based retrieval, and four retrieval baselines: FR~\cite{lin2024flowretrieval}, STRAP~\cite{memmel2025strap}, BR~\cite{BehaviorRetrieval}, and IWR~\cite{importance}. We report the success rate (in \%) and standard deviation across 3 seeds. The best results are in \textbf{bold} and the second-best results are \underline{underlined}.}
    \label{tab:baselines_sim}
    \vspace{2mm}
    \centering
    \setlength{\tabcolsep}{1.9mm}
    \begin{tabular}{lccccccccccc}
    \toprule
    Category & \multicolumn{5}{c}{Pick-Place} & \multicolumn{2}{c}{Spatial Understanding} & \multicolumn{3}{c}{Composite Manipulation} \\
    \cmidrule(lr){2-6}\cmidrule(lr){7-8}\cmidrule(lr){9-11}
    Task & Soup-S & Cheese-B & Book & Soup-C & Moka-M & Mug-Mug & Mug-P & Stove-M & Bowl-C & Mug-M & Average\\
    \midrule    
    BC  & 2.0$\pm$2.0  & 18.0$\pm$3.5 & 20.7$\pm$4.2  & 3.3$\pm$1.2  & 0.0$\pm$0.0 & 0.0$\pm$0.0  & 12.7$\pm$3.1  & 68.0$\pm$5.3  & \underline{77.3$\pm$4.2} & 24.7$\pm$5.8 & 22.7$\pm$1.1 \\
    Random & 10.0$\pm$5.3  & 18.0$\pm$3.5 & 72.0$\pm$9.2  & 6.7$\pm$2.3  & 0.7$\pm$1.2 & 11.3$\pm$2.3 & 8.0$\pm$5.3  & 58.7$\pm$4.2  & 44.0$\pm$2.0 & 30.7$\pm$9.2 & 26.0$\pm$3.7 \\
    All Data  & 16.7$\pm$1.2  & 19.3$\pm$3.1 & 81.3$\pm$6.1  & 11.3$\pm$2.3  & 3.3$\pm$4.2 & 12.7$\pm$3.1 & 6.7$\pm$2.3  & 62.7$\pm$6.4  & 44.7$\pm$3.1 & 31.3$\pm$11.7 & 29.0$\pm$2.4 \\
    Language  & 16.7$\pm$8.1  & 20.0$\pm$5.3 & 88.7$\pm$5.0  & \bf 14.7$\pm$8.1  & 2.7$\pm$3.1 & 16.0$\pm$3.5 & 10.7$\pm$2.3 & 68.0$\pm$5.3  & 45.3$\pm$9.0 & 32.7$\pm$6.1 & 31.5$\pm$1.1 \\
    FR & 16.0$\pm$5.3  & 20.0$\pm$0.0 & 76.0$\pm$7.2  & 9.3$\pm$5.0  & 0.7$\pm$1.2 & 14.0$\pm$0.0 & 11.3$\pm$5.8  & 63.3$\pm$6.1  & 46.7$\pm$7.6 & 25.3$\pm$10.3 & 28.3$\pm$1.6 \\
    STRAP & 11.3$\pm$1.2  & \underline{32.0$\pm$4.0} & 68.0$\pm$7.2  & 12.0$\pm$3.5  & \underline{6.7$\pm$1.2} & \underline{17.3$\pm$1.2} & \underline{16.7$\pm$1.2}  & 57.3$\pm$4.6  & 58.0$\pm$5.3 & \underline{40.0$\pm$11.1} & 31.9$\pm$2.1 \\
    BR & \underline{17.3$\pm$4.2}  & \bf 36.7$\pm$4.6 & \bf 95.3$\pm$2.3  & 9.3$\pm$2.3  & 3.3$\pm$1.2 & \underline{17.3$\pm$2.3} & 8.7$\pm$2.3  & 64.0$\pm$3.5  & 50.7$\pm$9.2 & 29.3$\pm$2.3 & \underline{33.2$\pm$1.5} \\
    IWR & 14.7$\pm$3.1  & 29.3$\pm$6.4 & 88.7$\pm$3.1 & 11.3$\pm$3.1  & 2.7$\pm$3.1 & 14.7$\pm$4.2 & 11.3$\pm$1.2  & \underline{70.7$\pm$3.1}  & 54.7$\pm$11.7 & 27.3$\pm$1.2 & 32.5$\pm$1.0 \\
   \rowcolor{lightgray!20} \textbf{\method} & \bf 18.0$\pm$5.3  & \underline{32.0$\pm$9.2} & \underline{90.7$\pm$1.2}  & \underline{12.7$\pm$5.0}  & \bf 15.3$\pm$3.1 & \bf 18.0$\pm$3.5 & \bf 26.0$\pm$2.0  & \bf 73.3$\pm$4.2  & \bf 81.3$\pm$11.0 & \bf 67.3$\pm$18.1 & \bf 43.5$\pm$3.3 \\							
    \bottomrule
    \end{tabular}
\end{table*}

%% file: table/real.tex
\begin{table}[t]
    \caption{\textbf{Experimental Results on Real-robot Manipulation}. For each task, we report the success rate over 20 trials. We also report the overall success rate across all tasks. The best results are in \textbf{bold}.}
    \label{tab:real}
    \vspace{2mm}
    \centering
    \begin{tabular}{lccccc}
    \toprule
    Task & Drawer & Trashcan & Cup-Drawer & Cup-Tea & Overall\\
    \#Subtask  & 1 & 2 & 3 & 4 &  \\
    \midrule
    BC & $ 12/20 $  & $7/20$ & $5/20$ & $3/20$ & $27/80$ \\
    BR & $15/20$ & $3/20$ & $4/20$ & $5/20$ & $27/80$ \\
    IWR & $16/20$ & $4/20$ & $7/20$ & $4/20$ & $31/80$ \\
    \rowcolor{lightgray!20}\textbf{\method}  & $\mathbf{18/20}$ & $\mathbf{8/20}$ & $\mathbf{13/20}$ & $\mathbf{9/20}$ & $\mathbf{48/80}$ \\
    \bottomrule
    \end{tabular}
\end{table}

%% file: table/ablation.tex
\begin{table}[t]
\begin{minipage}[t]{.5\linewidth}
\captionof{figure}{Performance under \textbf{Varying Reranking Retention Ratios}.}
\label{fig:ablation:percentage}
\centering
{\includegraphics[width=0.9\linewidth]{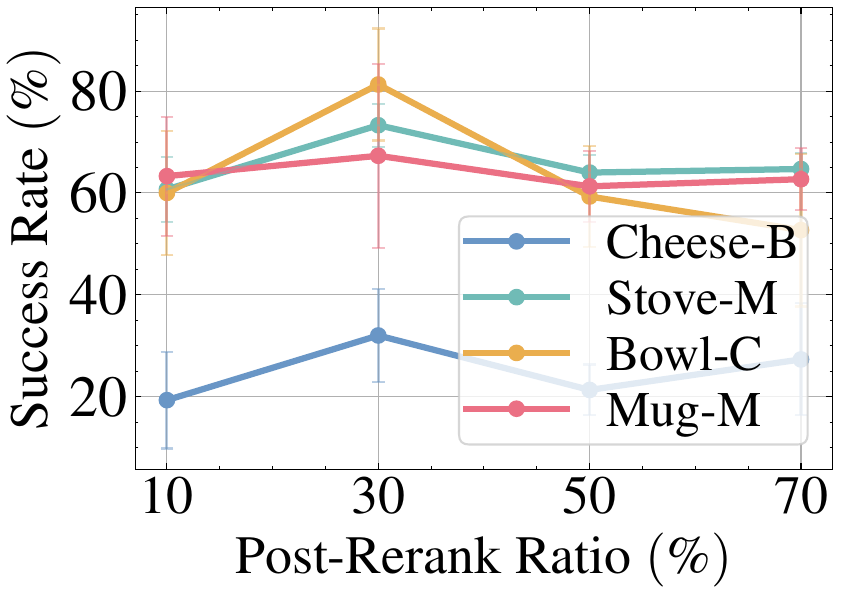}}
\end{minipage}\hspace{2mm}
\begin{minipage}[t]{.4\linewidth}
\caption{\textbf{Ablation on Task Decomposition}.}
\label{tab:ablation:decomposition}
\vspace{1mm}
\setlength{\tabcolsep}{0.5mm}{\begin{tabular}{lc}
\toprule
Method  & Avg. Succ. (\%) \\
\midrule
LLM Only & 37.1$\pm$2.5 \\
LLM+Skill Set & 37.7$\pm$1.2 \\
HSR & 43.5$\pm$3.3 \\
Human & 44.9$\pm$1.0 \\
\bottomrule
\end{tabular}}
\end{minipage}
\end{table}